\documentclass{article} 
\usepackage{iclr2026_conference,times}

\usepackage{amsmath,amsfonts,bm}

\def\eqref#1{equation~\ref{#1}}

\def\1{\bm{1}}

\DeclareMathAlphabet{\mathsfit}{\encodingdefault}{\sfdefault}{m}{sl}
\SetMathAlphabet{\mathsfit}{bold}{\encodingdefault}{\sfdefault}{bx}{n}

\usepackage{hyperref}
\usepackage{url}
\usepackage{wrapfig}
\usepackage{graphicx}
\graphicspath{ {./figures/} }
\usepackage{subcaption}
\usepackage{caption}
\usepackage{booktabs}

\usepackage{authblk}

\title{Large-Scale Chatbot Validation Through Customer Digital Twin Simulations}

\author{Cristovao Iglesias}
\author{Devesh Batra}
\author{Alankar Atreya}
\author{Stefan Wagner}
\author{Robert Hankache}
\author{Patrick Sinclair}
\author{Giulio Pelosio}
\author{Michael McMillan}
\author{Greig A.\ Cowan}
\author{Raad Khraishi}

\affil{NatWest AI Research}

\iclrfinalcopy 
\begin{document}

\maketitle

   \begin{abstract}
      LLM-based chatbots are transforming customer service in regulated domains such as banking, but scalable and cost-effective validation remains a critical barrier to safe deployment.
      We present a two-part contribution for large-scale chatbot validation.
      First, we introduce a methodology for creating high-fidelity synthetic customer agents (SCAs) as digital twins, grounded in real transactional and conversational data, that enables automatic generation and behavioral conditioning to simulate diverse customer profiles and interaction styles.
      Evaluation demonstrates that SCAs achieve high semantic alignment with real customers, low hallucination rates, and successful personality trait reproduction with controllable interventions.
      Second, we develop an SCA-based validation framework combining automated LLM-as-a-Judge evaluation, human expert testing, and adversarial probing.
      Scenario-based validation across emotional states, demographic groups, and linguistic factors confirms robust performance.
      Our approach was used to validate a customer facing chatbot at a leading UK bank, providing financial institutions with a scalable pathway toward regulatory compliance.
      \end{abstract}

\section{Introduction}

Large Language Model (LLM)-driven chatbots are reshaping customer service, delivering more flexible and scalable interactions than traditional interactive voice response (IVR) systems or human agents. In financial services, these chatbots now manage complex queries, allowing specialists to concentrate on high-value tasks. However, deployment in regulated environments raises unique validation challenges: chatbots must consistently achieve high accuracy, preserve conversational fidelity, ensure fairness across diverse users, resist adversarial manipulation, and adhere to stringent regulatory standards~\citep{fca2024aiupdate,nist2023airmf,zou2023universal}.

\noindent\textbf{Key challenge:} Traditional validation approaches, relying heavily on small-scale human trials and manual review, are costly, slow, and ill-suited for the iterative prompt engineering needed for LLM chatbot optimisation~\citep{deriu2021survey,reynolds2021promptprogramming}. They struggle to scale to the diverse edge cases and adversarial scenarios necessary for safe deployment in high-stakes financial contexts, creating a critical bottleneck that hinders timely deployment and continuous improvement~\citep{perez-etal-2022-red,zou2023universal}.

\textbf{Related work:} User simulation has long enabled scalable evaluation of task-oriented dialogue systems, from early frameworks such as PARADISE~\citep{walker-etal-1997-paradise} to agenda-based simulators~\citep{schatzmann-etal-2007-agenda}. Recent advances leverage large language models as higher-fidelity simulated users and evaluators, including LLM-based user simulation~\citep{davidson-etal-2023-user-simulation-llm}, dual-LLM verification~\citep{luo-etal-2024-duetsim}, and LLM-as-a-judge evaluation paradigms~\citep{zheng-etal-2023-judging,liu-etal-2023-g}. Our work extends these ideas to regulated financial services by grounding synthetic customers in real transaction context and enabling scalable, compliance-aware validation of banking chatbots.


\textbf{Contributions:} This paper presents:
1) A methodology for creating high-fidelity synthetic customer agents (SCAs) as digital twins, grounded in real transactional and conversational data. We also provide evidence that SCAs closely replicate real customer interactions in semantic content and personality traits, with low rates of hallucination and factual errors; and
2) A scalable SCA-based framework for large-scale chatbot validation.





\section{Synthetic Customer Methodology}


\begin{figure}[!t]
   \centering
   \includegraphics[width=\linewidth]{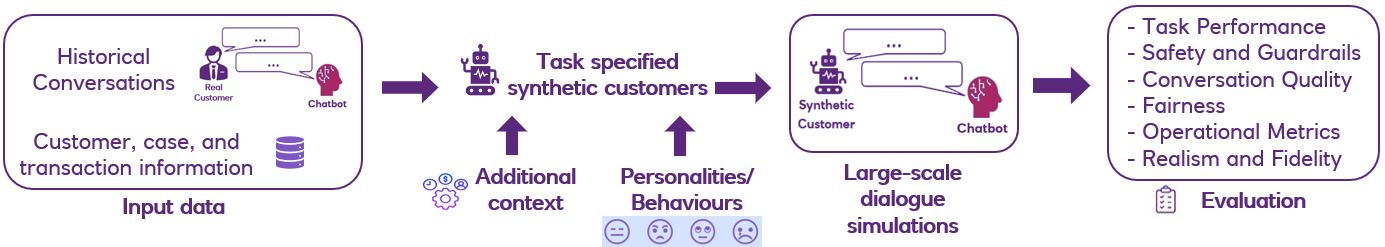}
   \caption{Overview of the synthetic customer agent (SCA) methodology.}
   \label{fig:sca_overview}
\end{figure}

We develop SCAs as LLM-powered digital twins that replicate real customer behavior in multi-turn conversations with chatbots. Our prompt engineering methodology supports two primary use cases: (1) \textit{transcript-driven simulation}, which faithfully mimics customers based on historical conversation transcripts and contextual data, and (2) \textit{personality-conditioned simulation}, which augments transcript-based behavior with specific emotional states or interaction styles.
Figure~\ref{fig:sca_overview} illustrates our end-to-end methodology. We begin with real-world input data (historical customer-agent conversations, case context, and transaction details) which are processed into general structured system prompt that configure each SCA. These prompts combine a general system scaffold with case-specific context and optional personality or behavioral interventions. The resulting SCAs engage in large-scale, multi-turn dialogue simulations with the target chatbot, enabling systematic exploration of diverse customer profiles and interaction patterns under realistic operational conditions.

\textbf{Formal Definition.} For a given real customer, we define the digital twin $g$ as an LLM-based agent that generates responses $r_t = g(q_t, s, h, p)$ at each conversational turn $t$, where:
$q_t$ is the chatbot's question or prompt at turn $t$.
$s$ is the scaffolding: task instructions ensuring fidelity to real customer behavior, including example phrases grounding responses in authentic human language.
$h$ represents historical context: the customer's past interactions, case details, and transaction information.
$p$ denotes personality interventions: optional behavioral modifiers (e.g., angry, anxious, confused) that adjust the SCA's emotional tone and interaction style.
This formulation enables controlled experimentation: by varying $p$ while holding $h$ constant, we can generate counterfactual scenarios to test chatbot robustness across diverse customer behaviors. By conditioning on real historical data $h$, we ensure that SCAs remain grounded in authentic customer contexts.

The conversations generated through SCA-chatbot interactions are evaluated across multiple dimensions (task performance, safety and guardrail compliance, conversation quality, fairness, and realism) producing structured metrics and diagnostics. These outputs support downstream applications including task performance testing, red-teaming and guardrail validation, behavioral robustness assessment, UX exploration, and fairness screening, enabling scalable, reproducible, and policy-aligned validation of conversational AI systems in regulated environments.

\subsection{Evaluation of SCA}

\textbf{Assessing Semantic and Lexical Similarity.}
%
We ran a transcript-driven multi-turn realism experiment on 600 anonymised transcripts (validation set) with sensitive data (e.g., PII, PAN) masked to guarantee privacy and regulatory compliance. For each simulation, we prompted a synthetic customer with each real chatbot assistant turn to generate 30 responses, then compared the concatenated synthetic customer message sequences against their real counterparts using cosine similarity (semantic meaning) \citep{steck2024cosine} and BLEU (word/phrase overlap) \citep{papineni2002bleu}.
Results show high semantic alignment but low lexical overlap with the best performance achieved by using GPT‑4.1 (T=0) as the base model, with negligible across temperature values, see Table \ref{tab:realism-metrics}. In short, synthetic responses are meaningfully similar to real ones (capturing the same main points and intent) while typically using different words and phrasing, yielding semantically strong but lexically diverse outputs.

\textbf{Evaluating Factual Faithfulness and Error Impact.}
We evaluated the fidelity of synthetic conversations against their real counterparts using a 750 anonymised transcripts (test set).
An LLM‑as‑Judge assessed each transcript on three binary metrics: conversation completeness, information wrongness, and inventing facts, enabling us to attribute misclassification causes.
Results show that synthetic conversations generally preserve the factual content of real transcripts, with fidelity issues occurring infrequently.
Most deviations involve omissions of contextual or behavioural details, minor inaccuracies (e.g., small amount differences), or occasional invention of new details (e.g., extra notifications), see Table \ref{tab:fidelity-misclassification}.
Omissions were the primary cause of misclassifications (11 cases), followed by invented facts (7 cases); inaccuracies almost always co-occurred with these.
The overall misclassification (hallucination, i.e., fabricated information not grounded in the source data) rate due to fidelity issues was 3.2\% (24 out of 750), indicating a low impact on classification accuracy. These rare errors can be reliably identified with LLM-as-Judge evaluation, supporting effective detection and remediation, and confirming the value of synthetic customers as faithful proxies for model validation.

\begin{table}[t]
   \centering
   \begin{minipage}[t]{0.48\linewidth}
   \centering
   \footnotesize
   \setlength{\tabcolsep}{4pt}
   \renewcommand{\arraystretch}{1.1}
   \captionof{table}{Realism metrics for synthetic customer responses: cosine similarity (semantic) and BLEU (lexical) across SCA base models and temperatures (T).}
   \label{tab:realism-metrics}
   \begin{tabular}{@{}lccc@{}}
   \toprule
   \bf SCA model & \bf Cosine & \bf BLEU & \bf T \\
   \midrule
   GPT-4.1 nano & 0.810$\pm$0.050 & 0.150$\pm$0.060 & 1 \\
   GPT-4.1      & 0.849$\pm$0.040 & 0.170$\pm$0.060 & 1 \\
   GPT-4.1      & 0.852$\pm$0.040 & 0.190$\pm$0.070 & 0 \\
   \bottomrule
   \end{tabular}
   \end{minipage}\hfill
   \begin{minipage}[t]{0.48\linewidth}
   \centering
   \footnotesize
   \setlength{\tabcolsep}{4pt}
   \renewcommand{\arraystretch}{1.1}
   \captionof{table}{Misclassification attribution by fidelity error type across 750 cases.}
   \label{tab:fidelity-misclassification}
   \begin{tabular}{@{}lcc@{}}
   \toprule
   \bf Fidelity metric & \bf Count & \bf Percentage \\
   \midrule
   Conversation completeness & 11 & 1.46\% \\
   Information wrongness     & 6  & 0.80\% \\
   Inventing facts           & 7  & 0.93\% \\
   \midrule
   \textbf{Total}            & \textbf{24} & \textbf{3.20\%} \\
   \bottomrule
   \end{tabular}
   \end{minipage}
\end{table}

\begin{figure}[h]
   \centering
   \begin{subfigure}{0.45\textwidth}
       \centering
       \includegraphics[width=\textwidth]{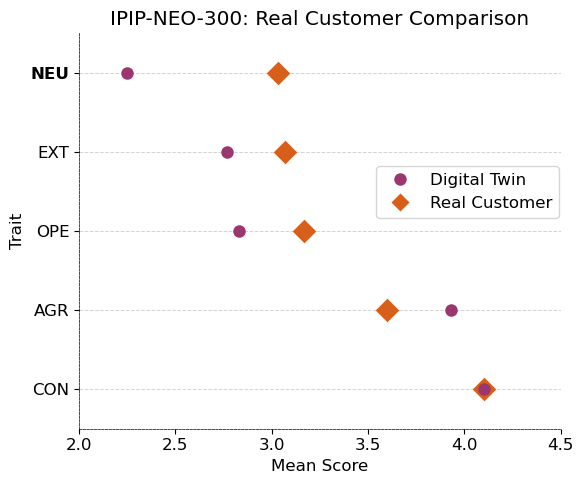}
   \end{subfigure}
   \hspace{0.05\textwidth} 
   \begin{subfigure}{0.45\textwidth}
       \centering
       \includegraphics[width=\textwidth]{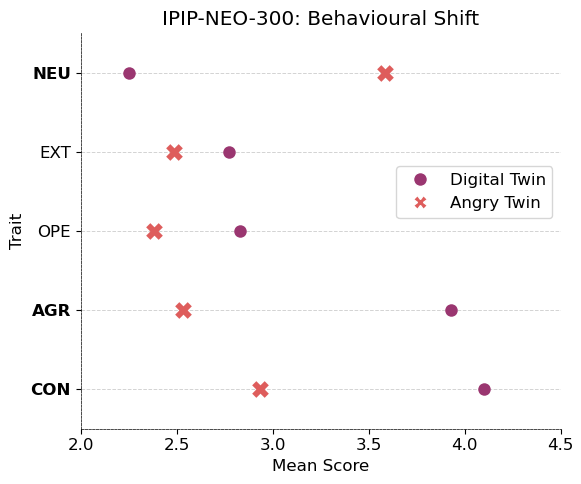}
   \end{subfigure}
   \caption{(Left) The digital twin does not differ from the real customer other than lower neurotiscism (NEU). (Right) We also show the effectiveness of interventions
    where the angry digital twin exhibits the personality traits for angry behaviour. The scale is from 1 (very inaccurate) to 5 (very accurate).}
   \vspace{-0.1cm}
   \label{fig:personality}
\end{figure}

\textbf{Personality Trait Modulation and Behavioural Realism.}
We assess personality alignment by analysing Big Five personality traits \citep{serapio2023personality, hartley-etal-2025-personality} in digital twin and real customer responses. An LLM rates how well each response set $R = {(r_{t})}_{t=1}^{T}$ aligns with predefined trait statements $E$ (scale 1-5). To test behavioural modulation, we apply personality markers to shift the digital twin towards 'angry' behaviour.
%
%
Figure \ref{fig:personality} (left) presents IPIP-NEO-300 scores for the Big Five personality dimensions (Neuroticism, Extraversion, Openness, Agreeableness, and Conscientiousness).
The digital twin closely matches the original scores (within 0.5 points on the 1–5 scale) across all dimensions except neuroticism.
This is unsurprising, as research on customer complaint dialogues often reveals higher levels of frustration and urgency \citep{matilla2002emotions, mcoll2009rage}. In contrast, (LLM)-based digital twins naturally tend towards polite behaviour. However, we demonstrate that targeted interventions can adjust the digital twin’s behaviour to reflect increased frustration.
As shown in Figure \ref{fig:personality} (right), conditioning the digital twin to exhibit angry behaviour leads to significant changes in IPIP-NEO-300 scores, particularly with higher neuroticism and lower agreeableness and conscientiousness, which are consistent with angry responses.
These results highlight a key strength of the digital twin approach: enabling controlled interventions on historical data to generate meaningful counterfactual scenarios.

\section{Framework for Validating a Chatbot Using SCA}




Leveraging the SCA methodology, we developed a validation framework for large-scale chatbot testing at a leading UK bank. Due to the sensitive nature of the specific application, we provide a high-level overview


\subsection{Chatbot Assessments}

The framework employs three complementary assessment methods to balance scalability with depth of insight:
\textbf{i) Automated evaluation (Auto-Eval):} We implement LLM-as-a-Judge to score the simulated conversations across nine dimensions: factual accuracy, summary accuracy, relevance, compliance, acknowledgement, language ease, smoothness, empathy, satisfaction, and frustration. Auto-Eval demonstrates strong agreement with subject matter expert (SME) ratings on objective metrics, enabling cost-effective continuous monitoring at scale.
\textbf{ii) Human expert testing:} SMEs conduct controlled testing by interacting with the chatbot under realistic scenarios, evaluating both classification accuracy and conversational quality. This human-in-the-loop validation provides ground truth labels and ensures the chatbot meets real-world usability standards.
\textbf{iii) Adversarial and safety testing:} Red-teaming exercises probe resilience to prompt injection, manipulation attempts, and inappropriate content. Complementary green-teaming identifies false positives where legitimate queries are incorrectly blocked. Automated guardrail pipelines balance safety with user experience.

\subsection{Chatbot Performance Metrics}
The framework tracks multiple metric categories to provide holistic performance assessment.
\textbf{i) Task performance metrics} include classification accuracy, precision, recall, and F1-scores for classification task.
\textbf{ii) Conversation quality metrics} assess nine dimensions through Auto-Eval, with separate tracking for objective versus subjective dimensions.
\textbf{iii) Operational metrics} monitor conversation dynamics (turn counts, words per turn, duration), resource usage (token consumption, cost), and guardrail-induced latency.
\textbf{iv) Safety and compliance metrics} track guardrail trigger rates, false positive/negative rates for content filtering, attack success rates against adversarial inputs, and adherence to regulatory messaging standards. This comprehensive metric suite enables both pre-deployment validation and continuous post-deployment monitoring.

\subsection{Chatbot Validation Across Different Scenarios}

The framework enables systematic testing across diverse scenarios by conditioning SCAs on specific characteristics.
\textbf{i) Baseline performance:} Transcript-driven SCAs faithfully replicate historical customer interactions to establish accuracy benchmarks across different case types.
\textbf{ii) Behavioral robustness:} Testing across emotional states (angry, anxious, confused, frustrated, panicked, neutral) and communication styles (talkative vs. silent) reveals stable performance, with only confused behavior showing increased inconclusive classifications due to insufficient information provision.
\textbf{iii) Fairness assessment:} SCAs conditioned on protected characteristics (gender, age, nationality) and proficiency regimes (CEFR Levels A1, B1, C1) confirm consistent accuracy across groups, with dynamic regime feedback improving lower-proficiency performance.
\textbf{iv) Base model comparison:}
Compare chatbot performance across different base models using identical test sets. The framework supports rapid evaluation of model updates by replaying historical conversations through new model versions, enabling data-driven decisions on model deprecation and replacement.

\section{Conclusion}

We present a two-part contribution for scalable validation of conversational AI in  regulated domains.
First, we introduce a methodology for creating high-fidelity SCAs as digital twins. Evaluation demonstrates high semantic alignment, low hallucination rates, and successful personality trait reproduction.
Second, we develop a comprehensive validation framework based on SCAs combining automated LLM-as-a-Judge evaluation, human expert testing, and adversarial probing. Scenario-based validation across emotional states, demographic groups, and linguistic factors confirms robust performance with no disparities. This provides financial institutions a scalable pathway to validating chatbot performance that meets regulatory expectations, with applicability to other regulated sectors.

\section*{Acknowledgements}

We acknowledge Amanda Miglionico, Daniel Clifton, Darryl Fishwick, and Martin Perry for their
domain expertise and insightful discussions on business logic. We also thank Tomoko Komatsu for
her support with customer testing, as well as all the subject matter experts (SMEs) who participated
in multiple testing rounds and provided valuable feedback for model improvement. Finally, we
extend our appreciation to Graham Smith and the Fraud Prevention CoE team for their support and
for enabling this work.

\bibliography{iclr2026_conference}
\bibliographystyle{iclr2026_conference}

\end{document}